\documentclass[pdflatex,sn-mathphys-num]{sn-jnl}

\usepackage{textcomp}

\usepackage{lineno}
\usepackage{graphicx}
\usepackage{setspace}
\usepackage{tocloft}
\usepackage{color, soul}
\usepackage{bm}
\usepackage{algorithm}
\usepackage{algpseudocode}
\usepackage{threeparttable}
\usepackage{booktabs}
\usepackage{float}
\usepackage[numbers]{natbib}

\usepackage{subcaption}
\usepackage{url}
\usepackage{multirow}

\usepackage[table]{xcolor}
\usepackage{wrapfig}
\usepackage{pifont}

\usepackage{amsmath,amssymb,amsfonts}
\usepackage{mathrsfs}
\usepackage{manyfoot}

\usepackage{listings}

\theoremstyle{thmstyleone}%
\theoremstyle{thmstyletwo}%

\theoremstyle{thmstylethree}%

\begin{document}

\title[Article Title]{Human-inspired Global-to-Parallel Multi-scale Encoding for Lightweight Vision Models}


\author*[1,2]{\fnm{Wei} \sur{Xu}}\email{xuweijonathan@gmail.com}

\affil*[1]{\orgname{Qinghai Normal University},\orgaddress{\street{Haihu Rd.}, \city{Xining}, \postcode{810016}, \state{Qinghai}, \country{China}}}
\affil[2]{\orgname{Qinghai Provincial Key Laboratory of IoT},\orgaddress{\street{Haihu Rd.}, \city{Xining}, \postcode{810016}, \state{Qinghai}, \country{China}}}


\abstract{Lightweight vision networks have witnessed remarkable progress in recent years, yet achieving a satisfactory balance among parameter scale, computational overhead, and task performance remains difficult. Although many existing lightweight models manage to reduce computation considerably, they often do so at the expense of a substantial increase in parameter count (e.g., LSNet, MobileMamba), which still poses obstacles for deployment on resource-limited devices. In parallel, some studies attempt to draw inspiration from human visual perception, but their modeling tends to oversimplify the visual process, making it hard to reflect how perception truly operates. Revisiting the cooperative mechanism of the human visual system, we propose GPM (Global-to-Parallel Multi-scale Encoding). GPM first employs a Global Insight Generator (GIG) to extract holistic cues, and subsequently processes features of different scales through parallel branches: LSAE emphasizes mid-/large-scale semantic relations, while IRB (Inverted Residual Block) preserves fine-grained texture information, jointly enabling coherent representation of global and local features. As such, GPM conforms to two characteristic behaviors of human vision perceiving the whole before focusing on details, and maintaining broad contextual awareness even during local attention. Built upon GPM, we further develop the lightweight H-GPE network. Experiments on image classification, object detection, and semantic segmentation show that H-GPE achieves strong performance while maintaining a balanced footprint in both FLOPs and parameters, delivering a more favorable accuracy-efficiency trade-off compared with recent state-of-the-art lightweight models.}


\maketitle

\section{Introduction}
\label{intro}
In recent years, the demand for visual intelligence on mobile and edge devices has continued to rise, driving the community to explore lightweight neural networks\citep{mobilenets, mobilenetv2, mobilenetv3,emo,mehta2021mobilevit,mehta2022separable,wang2024repvit} that can retain strong performance under constrained computational and memory budgets. However, existing approaches often struggle to strike a desirable balance among computational cost, parameter scale, and accuracy, particularly when deployed on resource-limited platforms \cite{edgevit,edgenext,efficientformer}.

Lightweight convolutional models (such as StarNet \cite{starnet} and MobileNetV4 \cite{mobilenetv4}), significantly reduce computation through structural simplification and have demonstrated promising design insights. Yet, their recognition capability and representational power remain bounded. Lightweight Transformers \cite{fdvit,mehta2021mobilevit,mehta2022separable,mobilevitv3,edgenext,mobilevig} partially ease hardware dependency, but their self-attention \cite{vit} focuses mainly on modeling spatial relations between tokens while offering limited channel modeling capacity, making it difficult to maintain both semantic richness and efficiency in lightweight regimes. Meanwhile, recent architectures like LSNet \cite{lsnet} and MobileMamba \cite{mobilemamba} achieve remarkable FLOPs reduction at the cost of notably increased parameter counts, leaving the efficiency–accuracy trade-off unresolved.

This leads to a natural question: how can we further unlock the performance potential of lightweight networks under strict hardware budgets, without sacrificing representational richness ?

\begin{figure}[!t]
\centering
\includegraphics[width=4 in]{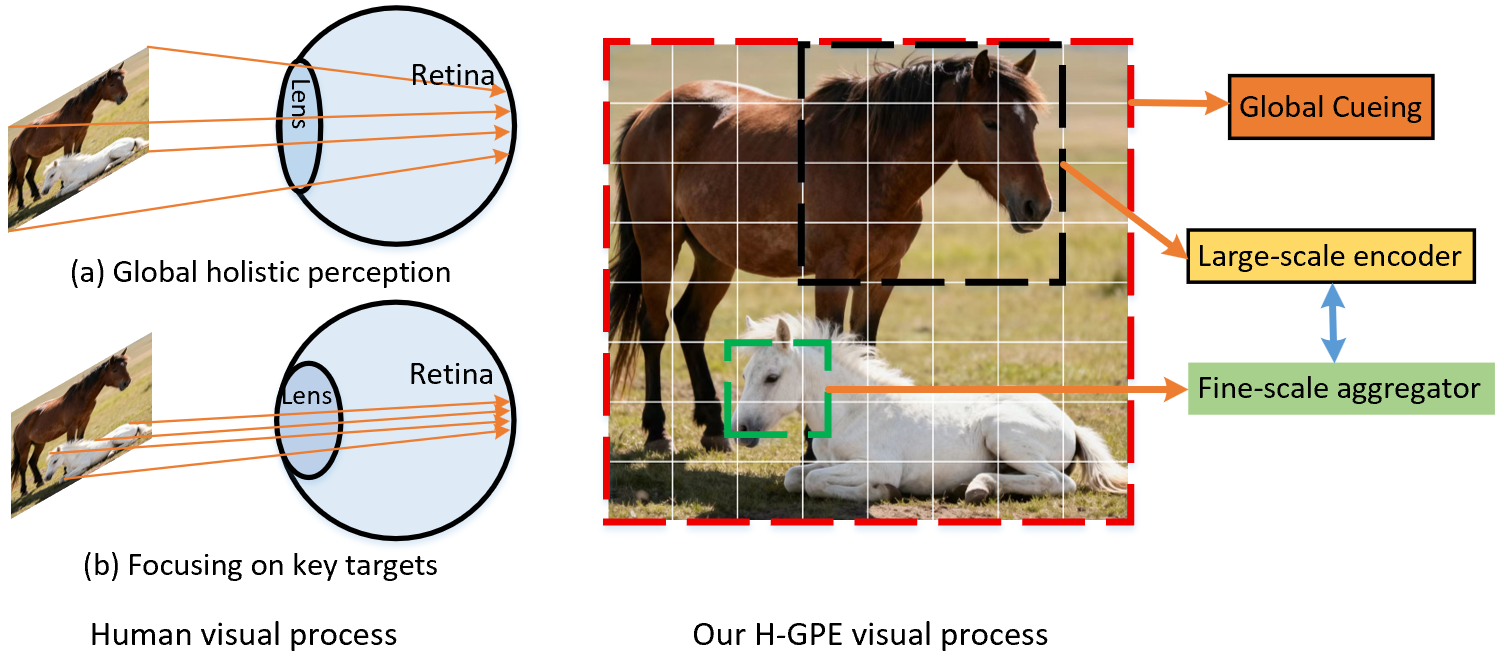}
\caption{Human visual processing pipeline and the recognition workflow of H-GPE.}
\label{visual}
\end{figure}

One promising direction is to revisit the principles of biological visual perception, which exhibit remarkable efficiency under limited neural and energy resources. Inspired by this observation, several studies attempt to design networks grounded in human visual mechanisms. For example, LSNet \cite{lsnet} draws inspiration from the dynamic multi-scale processing paradigm of “See Large → Focus Small”. While conceptually appealing, its implementation relies on a fixed 7×7 convolution to approximate large-field perception, which becomes increasingly inadequate for high-resolution feature maps. More importantly, many existing interpretations oversimplify the underlying physiological process, treating global perception and local focus as sequential or isolated operations, while overlooking their continuous interaction with peripheral and background awareness.

Based on a closer examination of the human visual system, we distill the core mechanism as follows: humans rapidly capture a coarse global layout and salient regions at a glance (global cue), and subsequently refine fine-grained details on focused targets, while persistently maintaining peripheral contextual awareness throughout the perception process \cite{human,human2}. This cooperative and parallel integration of global, local, and contextual information serves as the biological motivation for our design, as illustrated in Figure~\ref{visual}.

\begin{figure*}[!t]
\centering
\includegraphics[width=5 in]{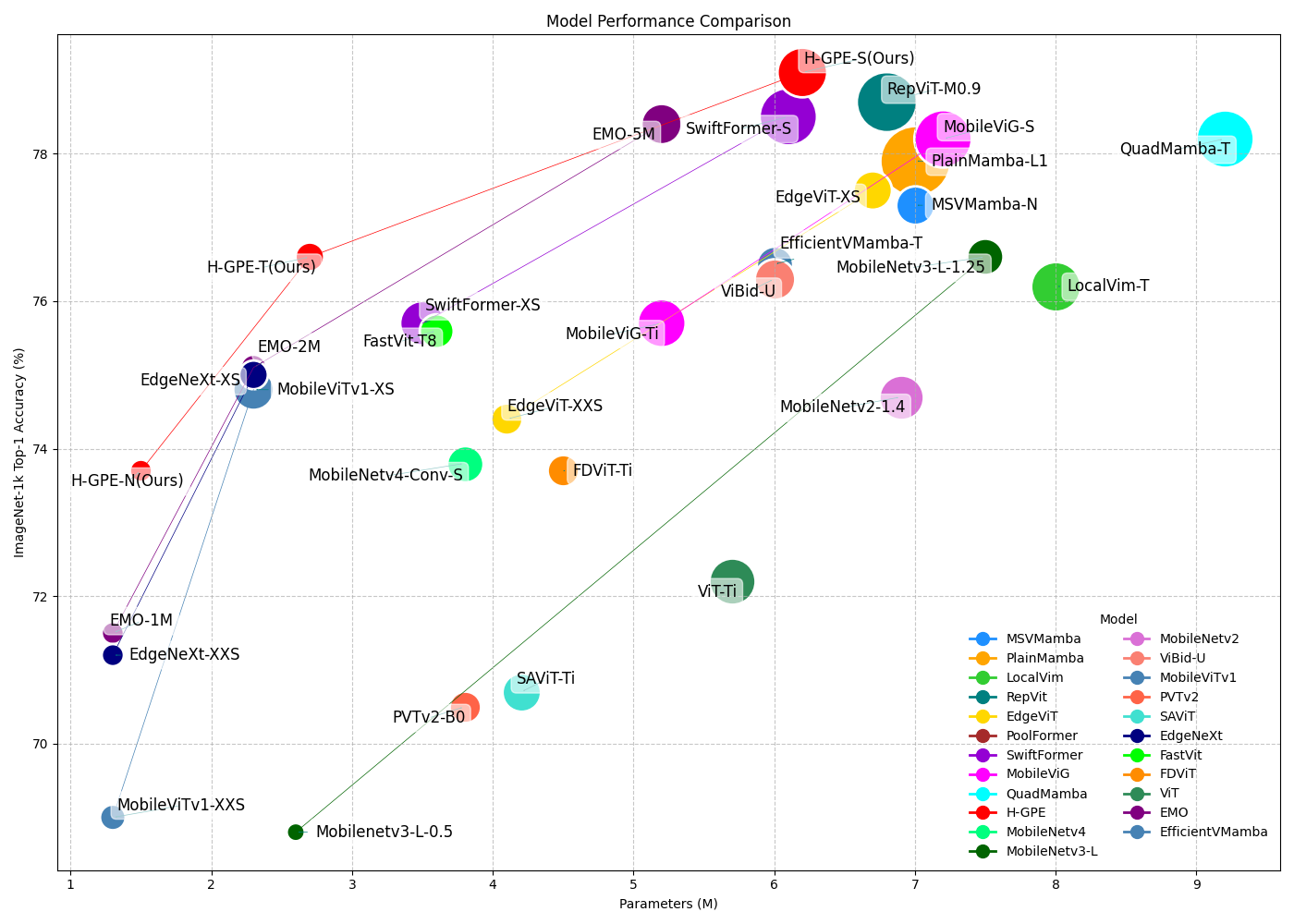}
\caption{Comparison of accuracy, parameter counts, and FLOPs on the ImageNet validation set, where the circle size encodes model complexity. H-GPE demonstrates strong competitiveness.}
\label{fig1}
\end{figure*}

Motivated by this understanding, we introduce GPM (Global-to-Parallel Multi-scale Encoding). At its core, the GIG (Global Insight Generator) performs global cueing, mimicking the human ability to rapidly estimate scene category and salient region distribution. GIG leverages strip pooling along with large - kernel grouped convolution to obtain global semantic priors in a lightweight form and injects them into subsequent branches. Following the principle of “global-first while retaining background during local focus”, the architecture executes two specialized branches in parallel along the channel dimension: LSAE (Large-Scale Attention Encoder) for mid-to-large scale semantic reasoning to maintain global coherence, and IRB \cite{mobilenetv2} for high-frequency detail aggregation to preserve texture, contour, and structural boundaries. To further enhance multi-scale complementarity, ASA (Axial Spatial Attention) is appended to the LSAE branch to strengthen directional sensitivity, whereas CRA (Channel Relational Attention) is applied to the IRB branch for optimal channel reweighting. From a frequency-domain perspective, GIG emphasizes low-frequency global information, LSAE (with ASA) targets mid-frequency semantic structures, and IRB (with CRA) captures high-frequency details, achieving multi-granularity decoupled modeling through collaborative encoding.

Building upon this design, we construct the lightweight network H-GPE (Human-inspired Global-to-Parallel Encoding), where GPE-Block is formed using IRB units with downsampling capability to build a complete backbone. On ImageNet classification \cite{deng2009imagenet}, H-GPE-S/T/N achieve 79.1\%, 76.6\%, and 73.7\% Top-1 accuracy with only 6.2M/1.5G, 2.6M/0.5G, and 1.5M/0.3G parameters and FLOPs, trained using a single NVIDIA RTX 4090 GPU. Furthermore, the model delivers superior performance on MS COCO \cite{2014coco} object detection and ADE20K  \cite{ade20k} semantic segmentation compared with existing lightweight counterparts, demonstrating a more favorable accuracy–efficiency balance, refer to Table \ref{tab6} and Figure \ref{fig1}.

The main contributions of this work are summarized as follows:
\begin{itemize}
    \item[1.]We propose the GPM architecture inspired by human cooperative vision, achieving multi-frequency decoupled modeling through global scanning → local focusing + background maintenance.
    \item[2.] We design H-GPE, a lightweight and scalable network applicable to classification, detection, and segmentation tasks.
    \item[3.] Extensive experiments show that H-GPE outperforms state-of-the-art lightweight models under comparable computational budgets, highlighting its potential for deployment in resource-constrained scenarios.
\end{itemize}

\section{Related Work}
\label{related work}
\subsection{Attention Mechanisms}
\label{related work:1}
\subsubsection{MHSA and Window-based MHSA}
Multi-head self-attention (MHSA) \cite{vit} lies at the center of Transformer-based vision models. By allocating several attention heads to process feature tokens in parallel, it effectively captures long-range relationships and strengthens feature representation. Vision Transformer (ViT) \cite{vit} demonstrated that splitting an image into patch tokens and applying MHSA enables the model to learn global interactions, achieving CNN-level or even superior performance when trained with sufficient data. DeiT \cite{deit} later refined this paradigm by employing aggressive augmentation and knowledge distillation, making training more data-efficient.

A well-known limitation of MHSA is its quadratic computational complexity $O(N^2)$ with respect to the number of tokens, causing the cost of computation and memory to escalate quickly as resolution increases. Swin Transformer \cite{liu2021swin} alleviates this issue by restricting attention computation to local windows and employing a shifted-window scheme to enable cross-window feature interaction. Building on this idea, PVT \cite{pvt} forms a multi-stage pyramid architecture to balance efficiency and representation ability, while works such as TNT \cite{tnt} and Shuffle Transformer \cite{shufflenet} investigate different token structures and window partitioning strategies to further refine performance. Nevertheless, when processing large images, WMHSA still demands considerable resources, and its deployment on resource-limited hardware remains challenging.

\subsubsection{Convolutional Attention Mechanisms}
Attention modules built on convolutional operations have been extensively adopted due to their lightweight nature and ability to emphasize informative regions. The Squeeze-and-Excitation block (SE) \cite{senet} recalibrates channel responses and initially sparked interest in channel attention research. ECA-Net (ECA) \cite{wang2020eca} simplifies SE by replacing fully connected layers with a 1D convolution, reducing parameters while maintaining effectiveness. FcaNet \cite{fcanet} introduces a frequency-aware perspective through DCT-based channel modeling. CBAM \cite{woo2018cbam} jointly applies channel and spatial attention, while SA-Net \cite{zhang2021sa} integrates both dimensions through channel permutation. Other variants, such as SFA \cite{sfa} and ELA \cite{ela}, enhance attention modeling by performing channel grouping or improving spatial filtering.

Overall, methods such as ECA and FcaNet emphasize channel aggregation, whereas CBAM, SA-Net, and SFA combine spatial and channel modeling. Nevertheless, compared with MHSA or WMHSA, convolutional attention mechanisms remain limited in capturing global dependencies, which restricts their effectiveness when used alone.

\subsection{Hybrid Transformer–CNN Models}
\label{related work:2}
Compared with MHSA or WMHSA, convolutional attention mechanisms often struggle to capture broad contextual interactions. For this reason, hybrid network designs combining convolution and attention have become increasingly common. MobileViT (v1/v2) \cite{mehta2021mobilevit,mehta2022mobilevitv2} and MobileFormer \cite{mobileformer} merge convolution with MHSA/WMHSA for mobile-friendly global-local reasoning. CMT \cite{cmt} embeds convolutional units inside Transformer blocks to reinforce local priors. Lite Transformer \cite{lite} mixes long- and short-range attention for richer feature encoding. EfficientFormer \cite{efficientformer} introduces resource-friendly WMHSA variants for real-time use, and EdgeViT \cite{edgevit} injects convolution into lightweight Transformers to enhance edge perception. Similarly EMO \cite{emo} fuse convolution with WMHSA to strike a balance between local detail preservation and global dependency modeling.

However, these hybrid designs share several limitations:
(1) attention modules often struggle to simultaneously model spatial structure and channel interactions effectively, hindering further reduction in parameters and FLOPs;
(2) implementation complexity remains high, with strong dependence on hardware capabilities;
(3) many models require large-scale data and strong augmentation pipelines to train stably;
(4) MHSA and WMHSA still emphasize spatial relations while offering limited channel modeling capacity.

In contrast, the GPM architecture introduced in H-GPE is inspired by the physiological dynamics of human vision—global scanning, focal refinement, and contextual maintenance-and integrates multiple specialized modules to collaboratively encode low-, mid-, and high-frequency information. This design leads to significant improvements in both performance and efficiency.

\section{The proposed method}

The proposed H-GPE is inspired by the human visual system, which first perceives the global scene and then focuses locally, while still maintaining awareness of the surrounding context. The GPM (Global-to-Parallel Multi-scale Encoding) architecture implements this principle by enabling collaborative modeling across low-, mid-, and high-frequency features. In this section, we describe the GPM design, followed by the construction of GPE-Blocks and the overall H-GPE network.

\subsection{GPM Architecture}

The GPM module begins with a Global Insight Generator (GIG, see Figure \ref{fig-attention}(a) and Table \ref{tab:GIG}) that performs panoramic feature retrieval, mimicking the human ability to grasp an overview of a scene in a single glance. GIG combines strip pooling \cite{hou2020strip} with large-kernel grouped convolution to efficiently capture global semantic and structural priors, which are then injected as global cues into subsequent branches.

Let the input feature map be $X \in \mathbb{R}^{C \times H \times W}$, where $C$ denotes the number of channels, and $H,W$represent the height and width, respectively. After GIG, the feature map is split along the channel dimension into two parallel branches.

and two specialized operators are applied in parallel:

\begin{figure}[!t]
\centering
\includegraphics[width=4 in]{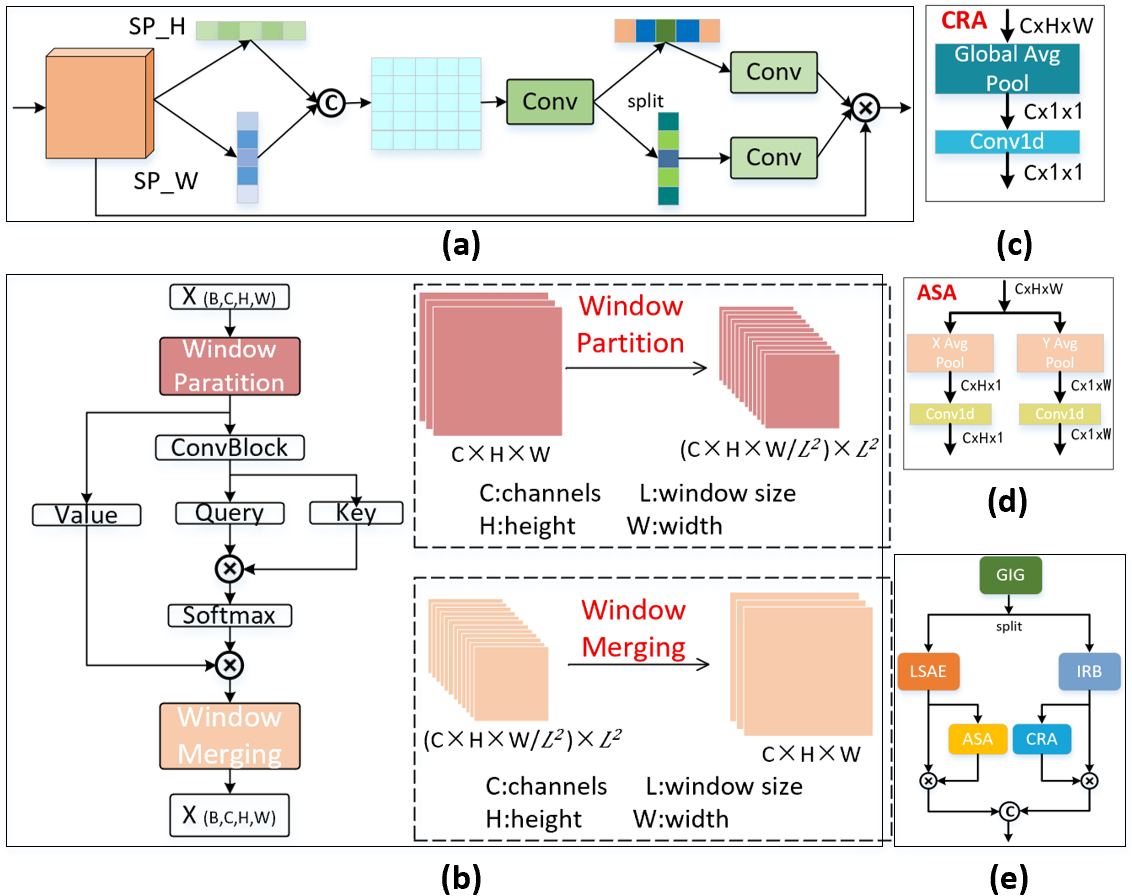}
\caption{(a) GIG module; (b) LSAE module; (c) CRA module; (d) ASA module; and (e) GPM architecture. 
$SP\_H$ and $SP\_W$ indicate strip pooling along the height and width directions, respectively;  $conv$ represents convolution; and Window Partition/Merging denote the splitting and merging of feature maps.}
\label{fig-attention}
\end{figure}

\begin{enumerate}
    \item LSAE (Large-Scale Attention Encoder, see Figure \ref{fig-attention}(b) and Table \ref{tab:LSAE} ): captures long-range dependencies over a large receptive field, ensuring semantic consistency and structural guidance across regions. ASA  is applied as an auxiliary module to enhance sensitivity to horizontal textures and vertical structures.
    \item IRB (Inverted Residual Block, see Figure \ref{fig-blocks}): performs fine-grained aggregation to strengthen textures, edges, and other high-frequency details. CRA attention is employed to optimize channel-wise feature weighting.
\end{enumerate}

Finally, the outputs of the two branches are concatenated to form the processed feature map. In the following, we describe GIG and LSAE (with ASA ) in detail and briefly introduce IRB \cite{mobilenetv2} (with CRA, see Figure \ref{fig-attention}(c)).

For panoramic feature retrieval, employing MHSA is considered unsuitable for lightweight networks. Instead, GIG uses strip pooling combined with convolution to obtain the global semantic and structural prior. Denote the full prior as $A_{all}$, the globally pooled representation as $y_{p}$, and the generated feature vector as $y\in \mathbb{R}^{C}$. The computation can be expressed as:

\begin{equation}
y_p = C(S_h, S_w),
\end{equation}

and the final global feature is obtained by

\begin{equation}
y = A_{\text{all}}\big( P_{\text{all}}(y_p), \text{G-Conv}_K \big),
\end{equation}

where $C(\cdot)$ denotes concatenation, $\text{G-Conv}_K$ is a grouped convolution with kernel size $K$ (we set $K=7$ to obtain a sufficiently large receptive field), and $S_h, S_w$ represent strip pooling along the height and width dimensions, respectively.

LSAE effectively models long-range dependencies while operating within local window partitions of the feature map. Let the window size be $w\times w$; based on empirical observations, we adopt $w=14$ or $w=7$ to capture mid- to large-scale contextual relations effectively. ASA attention is incorporated to enhance sensitivity to directional structures. 

The IRB module performs fine-scale aggregation using a $3\times 3$ convolution, following the original MobileNetV2 design. This design aligns with the small-kernel aggregation used in LSNet, ensuring high-frequency details such as edges and textures are preserved. CRA attention is applied afterward to optimize channel weighting across the feature map.

\subsubsection{GIG}

Based on the principle of panoramic retrieval and Eqs. (1)–(2), and drawing inspiration from the CA attention design in Hou et al.\cite{hou2020strip}, we note that a 1×1 convolution contributes little to receptive field expansion. Therefore, we consider that our design constitutes a more reasonable structure. The pseudocode of GIG is presented in Table \ref{tab:GIG}.

\begin{table}[ht]
\centering
\scriptsize 
\caption{Pseudocode for the GIG}
\begin{tabular}{l}
\toprule
\textbf{Require}: Input $X \in \mathbb{R}^{C \times H \times W}$ \\
\textbf{Ensure}: Output $y \in \mathbb{R}^{C \times H \times W}$ \\
1: $mip \gets \max(8, \lfloor C / \text{ratio} \rfloor)$ \\
2: $x_h \gets \text{AvgPool}(x, (H,1))$; $x_w \gets \text{AvgPool}(x, (1,W))^\top$ \\
3: $y \gets \text{Concat}(x_h, x_w, \text{dim}=2)$ \\
4: $y \gets \text{DWConv}(y)$; $y \gets \text{BN}(y)$; $y \gets \text{h\_swish}(y)$ \\
5: $x_h', x_w' \gets \text{Split}(y, [H,W], 2)$; $x_w' \gets x_w'^\top$ \\
6: $x_h' \gets \text{Sigmoid}(\text{DWConv}(x_h'))$; $x_w' \gets \text{Sigmoid}(\text{DWConv}(x_w'))$ \\
8: $Y \gets X \times x_h' \times x_w'$ \\
\textbf{Return}: $Y$ \\
\bottomrule
\end{tabular}
\label{tab:GIG}
\end{table}

The GIG module captures global structural and semantic cues by combining strip pooling with depthwise convolutions, generating attention maps that highlight informative regions while preserving overall context.

\subsubsection{LSAE and ASA}
After the GIG module, the feature map is split along the channel dimension into two branches, which can be expressed as:

\begin{equation}
Y \rightarrow \{ Y_0 \in \mathbb{R}^{\frac{C}{2} \times H \times W},\ Y_1 \in \mathbb{R}^{\frac{C}{2} \times H \times W} \}.
\end{equation}

The Large-Scale Attention Encoder (LSAE, see Figure \ref{fig-attention}(b)) applies self-attention within local windows to perform long-range modeling over a large receptive field. This ensures cross-region semantic consistency and shape cues, enhancing the model’s ability to accurately capture local features, analogous to how human vision perceives background information when focusing on a local region. The ASA attention acts as an auxiliary to LSAE, improving sensitivity to horizontal textures and vertical structures. Using local windows is a key design choice that keeps the module lightweight.

The design of LSAE is inspired by the MMB module proposed by Zhang et al. \cite{emo}. However, certain components such as SE attention contribute little in this context, and thus we simplify the structure. The pseudocode is designed following the workflow illustrated in Fig.\ref{fig-attention}(b), as presented in Table \ref{tab:LSAE}.

\begin{table}[ht]
\centering
\scriptsize 
\caption{Pseudocode for the LSAE}
\begin{tabular}{l}
\toprule
\textbf{Require}: Input feature $x \in \mathbb{R}^{C \times H \times W}$, denoted as $Y_0$ \\
\textbf{Ensure}: Output feature $y \in \mathbb{R}^{C \times H \times W}$, denoted as $\hat{Y}_0$ \\
1: $x \gets \text{Norm}(x)$ \\
2: \textbf{if} spatial attention enabled: \\
3: \quad Compute window size $(w_h, w_w)$; pad $x$ accordingly \\
4:\quad Reshape $x \to$ local windows of size $(w_h, w_w)$ \\
5:\quad $qk \gets \text{ConvBlock}(x, out\_ch=2C)$ \\
6:\quad Split $qk \to q, k$; $attn \gets \text{Softmax}(q k^\top / \sqrt{d})$ \\
7:\quad $v \gets \text{ConvBlock}(x)$; reshape $v \to$ windows \\
8:\quad $x \gets attn \cdot v$; reshape and remove padding \\
9: \textbf{else}: $y \gets \text{ConvBlock}(x)$ \\
\textbf{Return}: $y$ \\
\bottomrule
\end{tabular}
\label{tab:LSAE}
\end{table}

Here, according to Figure \ref{fig-attention}(b), \textit{Window Partition} and \textit{Window Merging} correspond to splitting the feature map into multiple local windows and recombining the processed windows back into a complete feature map, respectively. 
In Table \ref{tab:LSAE}, the padding and reshaping operations ensure proper alignment of the windows and restore the features to their original spatial dimensions.

Since LSAE is based on self-attention, the ASA  enhances the prominence of key regions in $\hat{Y}_0$ and facilitates the extraction of deeper semantic information. ASA also employs strip pooling, and following the design illustrated in Fig.\ref{fig-attention}(d), its pseudocode is presented in Table \ref{tab:ASA}. 

\begin{table}[ht]
\centering
\scriptsize
\caption{Pseudocode for the ASA}
\begin{tabular}{l}
\toprule
\textbf{Require}: Input feature $x \in \mathbb{R}^{C \times H \times W}$, denoted as $\hat{Y}_0$  \\
\textbf{Ensure}: Output feature $y \in \mathbb{R}^{C \times H \times W}$ \\
1: $x_h \gets \text{Mean}(x, \text{dim}=3)$; reshape $x_h \to (B, C, H)$ \\
2: $x_w \gets \text{Mean}(x, \text{dim}=2)$; reshape $x_w \to (B, C, W)$ \\
3: $x_h \gets \text{Sigmoid}(\text{GN}(\text{Conv1D}(x_h))))$; reshape $\to (B, C, H, 1)$ \\
4: $x_w \gets \text{Sigmoid}(\text{GN}(\text{Conv1D}(x_w))))$; reshape $\to (B, C, 1, W)$ \\
5: $y \gets x_h \cdot x_w$ \\
\textbf{Return}: $y$ \\
\bottomrule
\end{tabular}
\label{tab:ASA}
\end{table}

In the ASA module (as shown in Table \ref{tab:ASA}), we employ Group Normalization (GN) \cite{gn}. Compared with Batch Normalization, GN offers more stable training, higher performance, and better convergence.

\subsubsection{IRB and CRA}
\begin{figure}[!t]
\centering
\includegraphics[width=2 in]{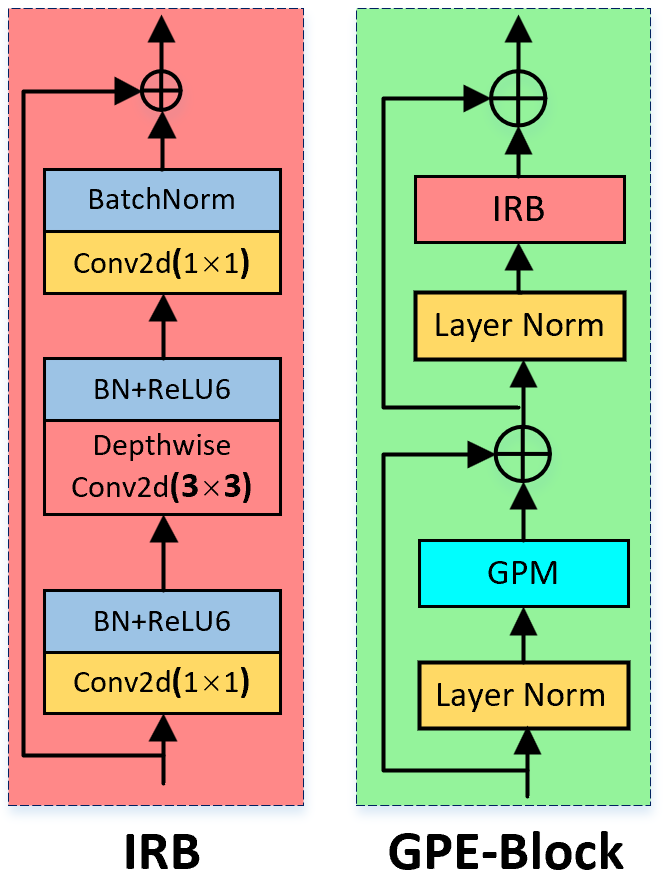}
\caption{IRB and GPM-Block,$BN$ denotes Batch Normalization}
\label{fig-blocks}
\end{figure}

The Inverted Residual Block (IRB) was first introduced in MobileNetV2. It is a residual-style module that first expands the channels via a pointwise projection convolution, then applies a depthwise convolution, and finally reduces the channels using another projection convolution. In our design, the IRB in the GPM structure, the GPE-Block, and in the Stem and Stage 1 of the H-GPE model does not reduce spatial resolution. However, in Stages 2–4 of H-GPE, the feature-map height and width are reduced to half of their original size. The IRB module is illustrated in Fig. \ref{fig-blocks} and Fig. \ref{fig-gpe}.

CRA (Channel Relational Attention, see Fig.\ref{fig-attention}(c)) is essentially a variant of ECA attention \cite{wang2020eca}, but its 1D convolution uses a kernel size determined by the cross-channel interaction range. Since this interaction range needs to be enlarged, we refer to this variant as CRA (see Fig. 5). We employ CRA to strengthen channel-wise weighting in the IRB branch features.

\subsubsection{Summary of GPM}
In this way, the complete GPM structure faithfully implements the visual-system principle of “global first, then local; and while focusing locally, still perceiving contextual background.” It jointly captures global and local semantic information, balances spatial and channel cues, and supports coordinated encoding of high-, mid-, and low-frequency components.

We compute the parameter count and FLOPs of the entire GPM, and compare them with ViT and Swin Transformer in Table 4. Here, the input/output feature maps lie in $\mathbb{R}^{C \times H \times W}$; $L = W \times H$, $l = w \times h$, where $(H, W)$ and $(h, w)$ denote the feature and window sizes, respectively, and $k$ denotes the convolution kernel size (including depthwise and $1 \times 1$ convolutions). With a window multi-head attention head dimension of $d$, the resulting parameter count and FLOPs of GPM are approximately $2C^{2}+C(3K^{2}+4K)+K$ and $O(C^{2}(2+\frac{3L}{2})+C(K^{2}L+6K^{2}L+16KH+\frac{9}{2}L)+Kd   )$, respectively. Compared with WMHSA, whose parameters are $4(C+1)C$ and FLOPs are $8C^{2}L + 4CLl + 3Ll$, the GPM is indeed extremely lightweight.

\subsection{GPE-Block}

The dual-residual structure of Transformers remains highly effective, and we retain it in H-GPE. Given the strong performance of the GPM module, it naturally serves as the core operator in the first residual branch. For the operator in the second residual branch, our choice follows the analysis of ConvNeXt \cite{convnet} as well as our own observations: spatial MLPs mainly perform channel transformations, whereas the Inverted Residual Block (IRB) not only provides channel manipulation but also offers stronger spatial representation capability while reducing FLOPs. Hence, IRB is a more suitable core operator for the second residual branch. For normalization, we continue using Layer Normalization \cite{ln}. The overall structure of the GPE-Block is illustrated on the right side of Fig.\ref{fig-attention}(e).

Since the GPE-Block does not rely on positional embeddings and its structure remains compact, its parameter count, FLOPs, and memory footprint are all very low. This efficiency is one of the reasons we can train the model to high accuracy using only a single GPU.

\subsection{H-GPE Model}
\subsubsection{Stem and Downsample Layers}

\begin{figure*}[!t]
\centering
\includegraphics[width=6 in]{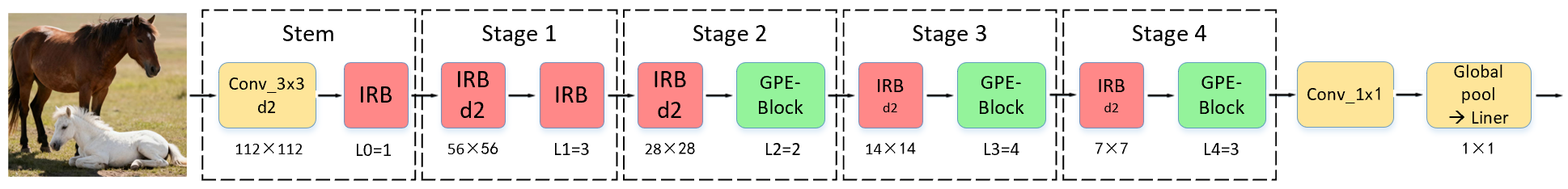}
\caption{Overview of the H-GPE architecture}
\label{fig-gpe}
\end{figure*}

Since the H-GPE model is designed as a lightweight backbone, the stem and downsample layers are also built with efficiency and low computational cost as core objectives. Owing to the strong performance of the IRB, we continue to adopt it in both the stem and downsample parts, as illustrated in Fig. \ref{fig-gpe}.

The stem is intentionally simple: we use only a single 2D convolution block (with a 3×3 kernel and stride 2), followed by an IRB module that preserves spatial resolution. As a stem module, this design already provides sufficient capability for handling low-level semantics.

The primary function of downsample layers is to reduce the spatial resolution of feature maps. We use IRB with stride set to 2 denoted as IRB d2 in Fig. \ref{fig-gpe} to further optimize the model.

\subsubsection{Stacking Rules}

After configuring the GPE-Block, and following the architectural patterns of MoConv \cite{moconv} and the design principles of EfficientNet \cite{efficientnet}, we construct H-GPE using a similar staging strategy.

The entire model consists of a Stem and four Stages, as shown in Fig. \ref{fig-gpe}. As previously described, both GPE-Block and IRB serve as basic blocks. Stage 1 uses IRB blocks without reducing resolution. In Stages 2–4, each basic block is composed of GPE-Block + IRB.

There are two key hyperparameters:

(1) Stack Count — the number of times the core block is repeated in each stage (IRB for Stage 1; GPE-Block for Stages 2–4).

(2) Out Channels — the output channel width of each stage.

The exact configurations are shown in Table \ref{tab4}. Additionally, the expansion ratio of the 3×3 convolution inside IRB differs across variants: the ratios are 6, 2, and 2 for H-GPE-S, H-GPE-T, and H-GPE-N, respectively.

Table \ref{tab4} summarizes the core hyperparameter settings for the H-GPE-S (Small), H-GPE-T (Tiny), and H-GPE-N (Nano) variants.

\begin{table}[htbp]
\centering
\scriptsize 
\caption{Key hyperparameters and performance metrics of H-GPE-S (Small), H-GPE-T (Tiny), and H-GPE-N (Nano). P represents parameters, and F denotes FLOPs.}\label{tab4}
\begin{tabular} {c | c | c | c | c  }
\hline
        Model             & Stack\_count         & Out\_channels        & Params(M)      & Flops(G) \\
         \hline
          H-GPE-S        & [3,2,4,3]           & [64,128,192,256]            & 5.6        & 1.4  \\

          H-GPE-T        & [3,2,4,3]           & [64,96,128,160]             & 2.3        & 0.5 \\

          H-GPE-N        & [3,2,4,3]           & [48,64,80,112]              & 1.2         & 0.3 \\

          \hline
\end{tabular}
\end{table}

\section{Experimental and Results}

In this section, we extensively evaluate the effectiveness of H-GPE across three fundamental vision tasks—image classification (ImageNet), object detection (MS COCO), and semantic segmentation (ADE20K). The results consistently demonstrate that, under comparable parameter counts and FLOPs, H-GPE surpasses a series of strong baselines, including state-of-the-art CNNs, hybrid CNN-Transformer architectures, and Mamba-based models. Notably, all models are trained on a single NVIDIA RTX 4090 GPU, underscoring the efficiency and practical deployability of H-GPE. In addition, a set of ablation studies further confirms the contribution and necessity of each key component in our architecture.

\subsection{Image Classification on ImageNet}
\textbf{Implementation Details}

For image classification, we conduct experiments on the ImageNet benchmark, training H-GPE from scratch. ImageNet remains one of the most authoritative datasets in classification, where Top-1 accuracy serves as the primary metric for evaluating model performance. The dataset contains 1.28M training images and 50K validation images.

We adopt the publicly available CVNets framework \cite{cvnets} for training. AdamW \cite{adamw} is used as the optimizer, paired with label-smoothing cross-entropy (smoothing = 0.1), L2 weight decay, multi-scale sampling, and exponential moving averaging of model weights (momentum = 0.00015). The learning rate follows a cosine-annealing schedule , warming up for the first 2,000 iterations, and decaying from 0.0015 to 0.0001 thereafter. All models are trained for 300 epochs with an input resolution of $224\times224$. Unless otherwise stated, the configuration aligns with MobileViTv1, and only basic data augmentation is applied—neither Mixup \cite{mixup} nor CutMix \cite{cutmix} is used.

\textbf{Results}

We evaluate three lightweight variants of H-GPE on the ImageNet validation set, with results summarized in Table 5. Remarkably, despite being trained under limited computational resources, our models achieve competitive and often superior performance.
\begin{enumerate}
\item H-GPE-N surpasses MobileViTv1-XXS by 4.7\% Top-1 accuracy, and outperforms EMO-1M by 2.2\%, demonstrating clear superiority at this scale.

\item H-GPE-T achieves higher accuracy than EMO-2M and MobileViTv1-XS, improving by 1.5\% and 1.8\%, respectively.

\item H-GPE-S outperforms SpectFormer-T by 2.2\%, while also showing advantages in both model size and accuracy.
\end{enumerate}

Overall, the results in Table 5 verify that H-GPE attains outstanding accuracy with exceptionally low parameters and computational cost, even on a large-scale benchmark like ImageNet. This strongly validates the effectiveness of designing vision models guided by the mechanisms of human visual perception.

\begin{table}[htbp]
\centering
\scriptsize 
\caption{Classification performance on ImageNet \cite{deng2009imagenet} dataset.  “Pub” indicates publication date and source.}\label{tab5}%
\begin{tabular}{ l | c | c | c| c }  
\toprule
         Model                                              & Params(M)                  & GFLOPs         & Top-1(\%)     &Pub	 \\
         \midrule

          PVT-ACmix-T\cite{pvt}	                       & 13.2	             & 2.0 	         & 78.0	        &CVPR'22       \\
          MPViT-T\cite{mpvit}		                       & 5.8            	 & 1.6           & 78.2	        &CVPR'22       \\
          ResMLP-S12 \cite{resmlp}	                       & 15          	     & 3.0	         &76.6          & PAMI'22     \\
          PoolFormer-s12\cite{metaformer}	               & 12  	             & 4.0           & 77.2	        & CVPR'22      \\

          ConvMLP-S	\cite{convmlp}	                       & 9	             & 4.8           &76.8	        & CVPR'23      \\
          RepViT-M0.9 \cite{wang2024repvit}               & 6.8	             & 2.2           &78.7	        & ICCV'23     \\
          SwiftFormer-S \cite{swiftformer}		         & 6.1	             & 2.0           & 78.5	        & ICCV'23      \\
          EMO-5M \cite{emo}		                        & \textbf{5.1}	     & 0.9            & 78.4         & ICCV'23       \\
          MobileViG-S \cite{mobilevig}		             & 7.2	             & 2.0           & 78.2         & CVPR'23       \\
          FLatten-PVT-T \cite{flatten}	                 & 12.2         	     & 2.0	         & 77.8	        & ICCV'23       \\
          StarNet-S4 \cite{starnet}                                    &7.5             &1.0             &78.4         & CVPR’24      \\
            
          QuadMamba-T \cite{quadmam}                    & 10                  & 2             & 78.2         & NIPS’24   \\
          localmamba-T \cite{localmamba}                 & 8              & 1.5             & 76.2         & ECCV'24    \\
          Vim-S\cite{vim}                               & 7.9              & -             & 76.1         & ICML’24   \\
          MobileMamba-S6 \cite{mobilemamba}                             & 15.0	             & 0.6          & 78.0         & CVPR’25      \\
          LSNet-S \cite{lsnet}                                       & 16.1	        & \textbf{0.5}    & 77.8         & CVPR’25      \\
          MambaOut-Femto \cite{mambaout}                               & 7	             & 2.4          & 78.9         & CVPR’25      \\
          MoConv-S \cite{moconv}                                     & 5.4	             & 1.4           & 78.6         & JRIP’25      \\
          EMB-S \cite{emb}                                       & 5.9	             & 1.5           & 78.9         & IVC’25      \\
          EVMamba-S\cite{effmamba}                     & 11                 & 1.3             & 78.7         & AAAI’25   \\
          SpectFormer-T \cite{spectformer}	           & 9.2	             & 1.8          & 76.9         & CVPR’25      \\
          
          H-GPE-S(Our)                                & 6.1	             & 1.5           & \textbf{79.1}   & -      \\
          \midrule

          MobileViTv1-XS \cite{mehta2021mobilevit}         &2.3                 &1.0           &74.8          &ICLR'22	 \\
          EdgeViT-XXS \cite{edgevit}  	                   & 4.1	             & 0.6 	         & 74.4	        &ECCV'22       \\
          PVTv2-B0 \cite{pvt}                             & 3.7	             & 0.6 	         & 70.5	        &CVM'22       \\
          MobileViG-Ti \cite{mobilevig}                     & 5.2	             & 1.4 	         & 75.7	        &CVPR'22       \\
          FDViT-Ti \cite{fdvit} 	                        & 4.5	             & 0.6           & 73.7        & ICCV'23      \\
          SwiftFormer-XS \cite{swiftformer}	               &3.5                 &1.2           &75.7          &ICCV'23	 \\
          FastVit-T8 \cite{fastvit}                      &3.6                 &0.7           &75.6           &ICCV'23	 \\
          EMO-2M \cite{emo}		                         & \textbf{2.3}     & 0.4          & 75.1         & ICCV'23       \\
          tiny-MOAT-0 \cite{moat}		                  & 3.4	             & 0.8           & 75.5         & ICLR'23      \\
          MobileNetv4-C-S \cite{mobilenetv4}        &3.8                &0.4           &73.8          &ECCV'24	 \\
          StarNet-S2 \cite{starnet}                                         &3.7             &0.5             &74.8         & CVPR’24      \\
          Vim-Ti\cite{vim}                                 & 7                & -             & 73.1         & ICML’24   \\
          MobileMamba-T4 \cite{mobilemamba}                                & 14.0	             & 0.4          & 76.1         & CVPR’25      \\
          SegMAN-TEncoder \cite{segman}                      & 3.5                & 0.6             & 76.2         & CVPR’25   \\
          LSNet-T \cite{lsnet}                                         & 11.4	          & \textbf{0.3}          & 74.9         & CVPR’25      \\
          MoConv-T \cite{moconv}                                     & 2.3	             & 0.5           & 75.7         & JRIP’25      \\
          EMB-T \cite{emb}                                          & 2.5	             & 0.6            & 76.3	        & IVC’25          \\
          H-GPE-T(Our)                                   & 2.6	             & 0.5           & \textbf{76.6}   & -      \\

          \midrule
          
          MobileViTv1-XXS \cite{mehta2021mobilevit}        & \textbf{1.3}         &0.4           &69.0         &ICLR'22	 \\
          SAViT-Ti \cite{savit}		                       & 4.2            	 & 0.9           & 70.7	        &NIPS'22       \\
          EdgeNeXt-XXS \cite{edgenext}		                   & 1.3            	 & 0.3 	         & 71.2     	&ECCVW'22    \\
          MobileViTv2-0.5 \cite{mehta2022mobilevitv2}        &1.4                 &0.4           &70.2          &arXiv'22	 \\
          EMO-1M \cite{emo}		                            & \textbf{1.3}	       & \textbf{0.3}           & 71.5         & ICCV'23       \\
          StarNet-S1 \cite{starnet}                                         &2.9                  &0.4                  &73.5         & CVPR’24      \\
          MoConv-S \cite{moconv}                                         & 1.2	             & \textbf{0.3}           & 72.2         & JRIP’25      \\
          EMB-N \cite{emb}                                           & 1.4              & \textbf{0.3}           &73.5	        & IVC’25          \\
          MobileMamba-T2  \cite{mobilemamba}                                  & 8.8	         & \textbf{0.3}          & 73.6         & CVPR’25      \\
          H-GPE-N(Our)                                   & 1.5	             & \textbf{0.3}           & \textbf{73.7}   & -      \\
          \bottomrule

\end{tabular}
\end{table}

\subsection{Object Detection}
\textbf{Implementation Details}

In this section, we evaluate the performance of H-GPE on object detection by using it as the backbone and integrating it with the SSDLite \cite{mobilenetv2} detection head. During training, input images are resized to 320×320, with a batch size of 64. We adopt AdamW as the optimizer and employ a cosine annealing strategy, where the initial learning rate is 0.0015 and the minimum learning rate is 0.00015. The model is trained for 200 epochs.

We load ImageNet-pretrained weights and conduct experiments on the MS COCO 2017 dataset  using the CVNets framework. All experiments are carried out on a single RTX 4090 GPU without the use of advanced data augmentation techniques such as Mixup, CutMix, etc.

The MS COCO (Microsoft Common Objects in Context) dataset is one of the most representative large-scale benchmarks for object detection, instance segmentation, keypoint detection, and image captioning, containing approximately 328K images. It is widely recognized as a core evaluation standard in the field.

\textbf{Results}

We compare H-GPE with several popular lightweight backbones of similar scale on SSDLite detection performance, as shown in Table 6. Our framework demonstrates a significant performance advantage over existing competitors.

Specifically, H-GPE-N achieves the highest accuracy with the lowest parameters and FLOPs, showing excellent efficiency. H-GPE-T also surpasses all models of similar size by more than 1.1\% in detection accuracy, while maintaining a lower computational cost. H-GPE-S further continues this trend and exhibits strong overall performance.

These results confirm that the bionic design philosophy of H-GPE is effective, and its global-perception-enhanced architecture provides strong generalization to downstream vision tasks such as object detection.

\begin{table}[htbp]
\centering
\footnotesize 
\caption{Object Detection  performance by SSDLite \cite{mobilenetv2} on MSCOCO2017 \cite{2014coco}.}\label{tab6}
    \begin{tabular}{ l | c | c | r  }  
\toprule
         Backbone                                      & Params(M)   & FLOPs(G)    & AP   \\
         \midrule
          ResNet50 \cite{he2016resnet}                  & 26.6         & 8.8      &25.2   \\
          MobileViTv1-S \cite{mehta2021mobilevit}       & 5.7         & 3.4      &27.7   \\
          MobileViTv2-1.25 \cite{mehta2022mobilevitv2}   & 8.2         & 4.7      &27.8   \\
          EdgeNeXt-S \cite{edgenext}                     & 6.2         & 2.1      &27.9   \\
          EMO-5M \cite{emo}                             & 6.0         & 1.8      &27.9   \\
          EMB-S \cite{emb}                                       & \textbf{5.6}     & 3.4      &\textbf{28.8}      \\
          PartialFormer-B1 \cite{partialformer}         & 8.0         & \textbf{1.5}      &27.1      \\
          H-GPE-S(Our)                                       & 6.1         & 3.2      &\textbf{30.3}      \\

          \midrule
          MobileViTv2-0.75 \cite{mehta2022mobilevitv2}    & 3.6	        & 1.8      & 24.6          \\
          EMO-2M \cite{emo}                               & 3.3         & \textbf{0.9}      &25.2   \\
          MoConv-S \cite{moconv}                                        & 3.7         & 1.6      &25.3      \\
          EMB-T \cite{emb}                                          & \textbf{2.7}    & 1.4      &25.7      \\
          PartialFormer-B0 \cite{partialformer}          & 5.0          & \textbf{0.9}      &24.3      \\
          H-GPE-T(Our)                                    & 2.9          & 1.3      &\textbf{26.8}      \\
          \midrule
          MobileNetv1 \cite{mobilenets}                   & 5.1	         & 1.3	     & 22.2          \\
          MobileNetv2 \cite{mobilenetv2}                  & 4.3	         & 0.8      & 22.1          \\
          MobileNetv3  \cite{mobilenetv3}                 & 5.0	         & \textbf{0.6}	     & 22.0          \\
          MobileViTv1-XXS \cite{mehta2021mobilevit}       & \textbf{1.7}	   & 0.9	     & 19.9          \\
          MobileViTv2-0.5 \cite{mehta2022mobilevitv2}	  & 2.0	         & 0.9	     & 21.2          \\
          EMO-1M \cite{emo}		                          & 2.3	         &\textbf{0.6}	    & 22.0           \\
          MoConv-T \cite{moconv}                                        & 2.5         & 1.0      &23.8      \\
          EMB-N \cite{emb}                                           & \textbf{1.7}	      & 0.9	     & 23.6             \\
          H-GPE-N(Our)                                   & 1.9	           & 0.8	     & \textbf{24.1}             \\
          \bottomrule

\end{tabular}
\end{table}

\subsection{Semantic segmentation}
\textbf{Implementation Details}

We further evaluate H-GPE on semantic segmentation using the ADE20K dataset. In this setting, we integrate H-GPE with DeepLabv3 \cite{2017deeplabv3} and conduct all experiments based on the CVNets framework. The segmentation models are initialized with ImageNet-1K pretrained weights.

Specifically, we adopt the ImageNet-pretrained H-GPE backbone combined with DeepLabv3 [5]. During training, input images are resized to 512×512, with a batch size of 8. We employ AdamW as the optimizer and apply an EMA strategy with momentum 0.00015. The learning rate follows a cosine schedule, starting at 0.001 and decaying to 0.0001, and the model is trained for 120 epochs in total.

Introduced by MIT in 2016, ADE20K is a widely recognized benchmark for open-scene semantic understanding. It contains approximately 25K complex scene images covering diverse indoor and outdoor environments. Each image includes on average 19.5 object instances spanning 10.5 semantic categories, making it a challenging dataset for segmentation evaluation.

\textbf{Results}

Table 7 presents results comparing H-GPE to existing lightweight backbones integrated with DeepLabv3. When balancing accuracy, parameters, and FLOPs, H-GPE consistently delivers the best trade-off across model scales. For instance, H-GPE-S achieves the highest mIoU in the medium-size group, surpassing EMO-5M by 2.7\% and PoolFormer by 3.2\%, highlighting a clear performance advantage. Similarly, H-GPE-T and H-GPE-N achieve strong accuracy under minimal computational cost, further demonstrating the effectiveness of our architecture in segmentation tasks.

\begin{table}[htbp]
\centering
\footnotesize 
\caption{Semantic Segmentation Performance Comparison on ADE20K Dataset  with DeepLabv3  model.}\label{tab7}
    \begin{tabular}{ l | c | c | r  }  
\toprule
         Backbone                           & Params(M)   & GFLOPs    & MIoU   \\
         \midrule
          PoolFormer \cite{metaformer}          & 15.7        & 31      &37.2   \\
          FastViT-SA12 \cite{fastvit}         & 14.1         & 29      &38.0   \\
          MoConv-S \cite{moconv}              &\textbf{8.5}  &7.7      &38.2   \\
          ResNet50 \cite{he2016resnet}         & 28.5        & 46      &36.7   \\
          EMO-5M \cite{emo}                  & 10.3         & \textbf{5.8}     &37.8   \\
          EMB-S \cite{emb}                    & 10.1        & 8.9        &39.8     \\
          H-GPE-S(Our)                         & 10.6       & 8.3        &\textbf{40.5}     \\
          \midrule
          
          EMO-2M \cite{emo}                 & 6.9	         & \textbf{3.5}	     & 35.3          \\
          MobileMamba-B4 \cite{mobilemamba}   & 23           & 4.7     &36.6   \\
          EMB-T \cite{emb}                    & 5.7	         & 4.8	     & \textbf{37.3}             \\
          MoConv-T \cite{moconv}               &\textbf{5.5}   &4.3      &35.5      \\
          H-GPE-T(Our)                         & 6.0          & 4.5        &36.9     \\
          \midrule
          
          EMO-1M \cite{emo}                 & 5.6	            & \textbf{2.4}	     & 33.5          \\
          H-GPE-N(Our)                      & \textbf{4.2}       & 3.2        &\textbf{34.2}     \\

          \bottomrule
\end{tabular}
\end{table}

Across all three vision tasks—classification, detection, and segmentation—H-GPE demonstrates strong generalization ability and exceptional efficiency. When considering accuracy relative to parameters and FLOPs, our models consistently outperform many lightweight baselines using only a single consumer-grade GPU. These results confirm the success of our design philosophy, emphasizing the value of human-vision-inspired modeling for efficient architectures.

\subsection{Ablation Experiments}
We further evaluate the contributions of GIG, LSAE, and ASA/CRA using H-GPE-N as the baseline. Each variant is trained for 100 epochs under identical settings. Results are summarized in Table \ref{tab8} and Table \ref{tab9}.

\begin{table}[htbp]
\centering
\footnotesize 
\caption{Classification performance on ImageNet-1K dataset. P (Params, M), T1 (Top-1, \%), F (GFLOPs). }\label{tab8}%
\begin{tabular}{ l | c | c | c |  r }  
\toprule
         Model Variant           &GIG               & P                  & F            & T1     	 \\
         \midrule
          H-GPE-N               & $\times$         & 1.4	             & 0.3           & 71.2       \\          
          H-GPE-N               & GIG $\rightarrow$ CA       & 1.5	             & 0.3           & 71.4      \\
          H-GPE-N              & GIG $\rightarrow$ ELA     & 1.4	             & 0.3           & 71.2      \\
          H-GPE-N               & GIG      & 1.5	             & 0.3           & \textbf{71.5}       \\

         \bottomrule

\end{tabular}
\end{table}

From Table \ref{tab8}, GIG delivers the most significant gain, indicating its effectiveness in introducing global scene awareness with minimal computational overhead. Unlike CA \cite{hou2021coordinate} and ELA \cite{ela}, which bring only marginal improvements, GIG provides a clearer global prior that better aligns with our human-vision-driven design philosophy. This suggests that early global cue extraction is crucial for lightweight backbones where receptive fields are inherently limited.

\begin{table}[htbp]
\centering
\footnotesize 
\caption{Classification performance on ImageNet-1K dataset. P (Params, M), T1 (Top-1, \%), F (GFLOPs). }\label{tab9}%
\begin{tabular}{ l | c | c | c | c | r }  
\toprule
         Model Variant           & LSAE             & ASA-CRA           & P             & F           & T1     	 \\
         \midrule     
          H-GPE-N                 & $\times$        &$\checkmark$       & 1.4	        & 0.3         & 71.2      \\
          H-GPE-N                 & $\checkmark$    & $\times$          & 1.5	        & 0.3         & 71.3      \\
          H-GPE-N                 & $\checkmark$    & $\checkmark$      & 1.5	        & 0.3         & \textbf{71.5}       \\
         \bottomrule

\end{tabular}
\end{table}

Table \ref{tab9} shows that LSAE contributes noticeably to representation learning by facilitating long-range dependency modeling within local windows, while ASA/CRA further refines the feature distribution through spatial semantic aggregation and channel-wise emphasis. When applied together, LSAE and ASA/CRA form a complementary pair: LSAE expands contextual perception, and ASA/CRA selectively reinforces informative responses. Their combination achieves the optimal performance of 71.5\% Top-1 accuracy without increasing FLOPs, underscoring the synergistic design of our multi-stage attention pipeline.

\section{Conclusion}
We introduce the human-vision-inspired GPM module and the lightweight H-GPE backbone. GIG provides a compact global prior, while two parallel branches—LSAE for mid/low-frequency semantics and IRB for high-frequency details—jointly refine representations, aided by ASA and CRA. With GPE-Block as the core, H-GPE-S/T/N achieve notable gains on ImageNet, MS COCO, and ADE20K, maintaining very low parameters and FLOPs under single-GPU training. This work operationalizes the principles of global-to-local perception and detail-focused processing with contextual awareness, enabling an efficient and scalable lightweight network that surpasses existing counterparts in accuracy–efficiency trade-offs.

Despite the strong results, several limitations remain: (1) GIG and LSAE rely on fixed kernel/window sizes, which may be suboptimal under varying resolutions or object scales—adaptive schemes will be explored. (2) The current design targets static images and lacks temporal modeling; extending GPM to video via lightweight temporal aggregation or frame attention is a natural next step. (3) Hardware-constrained evaluations limited us to FLOPs, parameters, and accuracy; future work includes latency/throughput benchmarking and further compression.

In summary, GPM and H-GPE provide an interpretable and efficient visual backbone through explicit multi-frequency decoupling. Future improvements-hardware -aware profiling, adaptive receptive fields, temporal extension, and structural compression—are expected to further enhance deployment in resource-limited scenarios.


\bibliography{ref}

\end{document}